# Situation-Aware Left-Turning Connected and Automated Vehicle Operation at Signalized Intersections

Sakib Mahmud Khan, and Mashrur Chowdhury, *Senior Member, IEEE*

*Abstract*— One challenging aspect of the Connected and Automated Vehicle (CAV) operation in mixed traffic is the development of a situation-awareness module for CAVs. While operating on public roads, CAVs need to assess their surroundings, especially the intentions of non-CAVs. Generally, CAVs demonstrate a defensive driving behavior, and CAVs expect other non-autonomous entities on the road will follow the traffic rules or common driving behavior. However, the presence of aggressive human drivers in the surrounding environment, who may not follow traffic rules and behave abruptly, can lead to serious safety consequences. In this paper, we have addressed the CAV and non-CAV interaction by evaluating a situation-awareness module for left-turning CAV operations in an urban area. Existing literature does not consider the intent of the following vehicle for a CAV's left-turning movement, and existing CAV controllers do not assess the following non-CAV's intents. Based on our simulation study, the situation-aware CAV controller module reduces up to 27% of the abrupt braking of the following non-CAVs for scenarios with different opposing through movement compared to the base scenario with the autonomous vehicle, without considering the following vehicle's intent. The analysis shows that the average travel time reductions for the opposite through traffic volumes of 600, 800, and 1000 vehicle/hour/lane are 58%, 52%, and 62%, respectively, for the aggressive human driver following the CAV if the following vehicle's intent is considered by a CAV in making a left turn at an intersection.

*Index Terms*—connected automated vehicle, autonomous vehicle, situation-aware, V2I, aggressive, rear-end

## I. INTRODUCTION

With the emergence of innovative computation and networking solutions, and novel sensor technology, Connected Automated Vehicle (CAV) will be mainstream in the future transportation system. However, CAVs will have to co-exist with the non-CAVs (i.e., human-driven vehicles) in the foreseeable future, and interacting with other surrounding objects for the shared roadway spaces can be challenging for CAVs [1], [2]. CAVs are operated by programmable controller software, and the logics embedded in the controller software are based on traffic rules and common driving norms/code of conduct. By default, CAVs are programmed to be 'defensive', which implies that the controllers are not allowed to violate any traffic rules. On the contrary, the driving behavior of different human drivers varies significantly. Based on the weather effects, and demography, psychology, and physical condition of the driver, humans can behave significantly different from each other. The driving behavior can also change due to the surrounding road conditions, and lack of available journey time [3]–[5]. In terms of aggressiveness, driver behavior can range anywhere from aggressive to non-aggressive and anything in-between. Dukes et al. (2001) classified aggressive drivers as active aggressive drivers (who behave abruptly) and passive-aggressive drivers (who induce others to act aggressively, e.g., driving slow and blocking others) [6]. Due to the aggressive nature of human drivers, a human can accelerate/decelerate abruptly, and maintain very little headway while following vehicles in front of them. This behavior often results in road rages or serious crashes. In urban areas, the presence of traffic signal controls could often lead to aggressive driving behavior [5]. The following aggressive driver can cause rear-end crashes if the leading vehicle suddenly decides not to cross the intersection and applies the brake. Also, if the front vehicle does not make any turn during the permissive phase, the following aggressive vehicle has to face a longer waiting time, and this can lead to road rage. The complicated interactions between CAVs (connected driverless vehicles with software making decisions based on input from in-vehicle sensors and wireless communication with the outside entities, such as other connected vehicles and roadside infrastructure) and non-CAVs (human-driven vehicles) result in varying driving behavior, which can lead to collisions in a mixed traffic scenario where both CAVs and non-CAVs coexist and share the same physical space. According to a real-world Autonomous Vehicle (AV) crash database, such conflicts between AVs and non-AVs exist [7]. Multiple studies found that the rear-end crash type dominates the total AV-related crash types in a mixed traffic scenario (i.e. traffic contains both AVs and non-AVs) [7]–[10]. For almost all of these rear-end crash cases, the primary reason was the following human driver applying poor braking and/or being distracted [7]. The conservative behavior of AVs is found to lead to potential conflicts with other non-AVs in the mixed traffic scenario [11]. The existing hierarchical planning architectures for AV controllers have a behavioral planning component where AVs decide about actions based on the

Manuscript received March 21, 2020.
Sakib Mahmud Khan is with Center for Connected Multimodal Mobility, Clemson University, SC 29631 USA (e-mail: sakibk@g.clemson.edu).

Mashrur Chowdhury is the Glenn Department of Civil Engineering, Clemson University, Clemson, SC 29634 USA (e-mail: mac@clemson.edu).

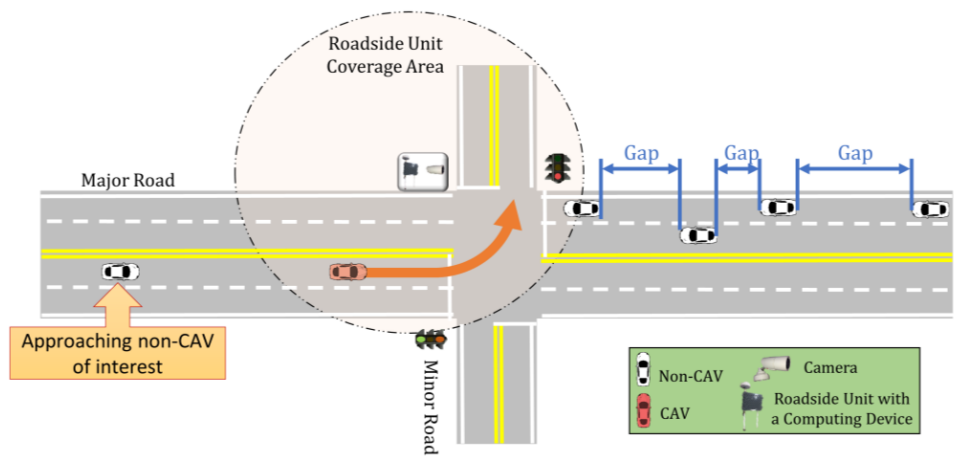

(a) In a TCPS based intersection

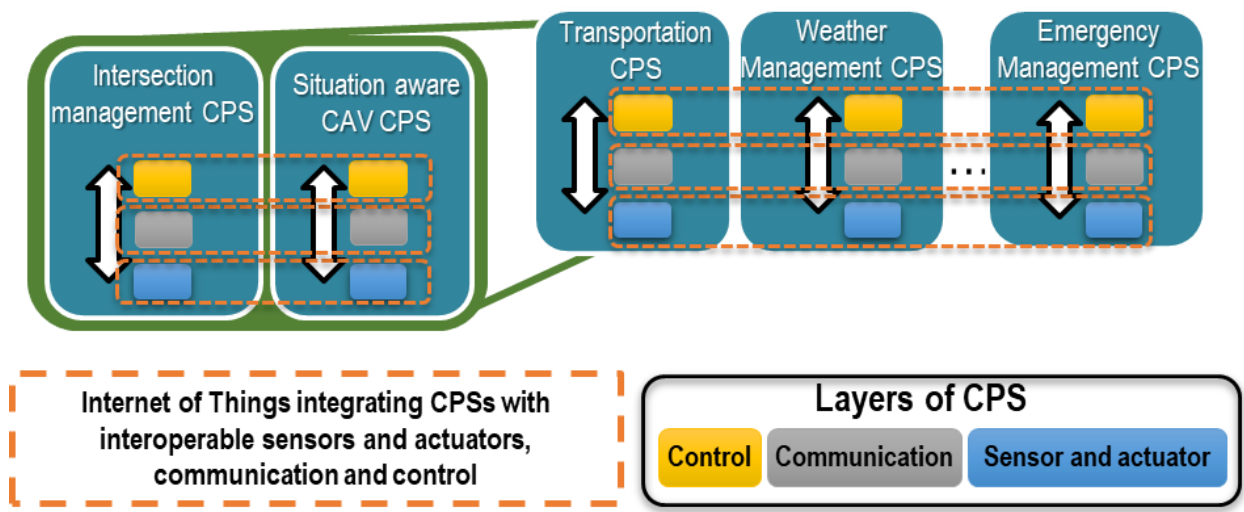

(b) In an IoT environment

Fig. 1. Situation-aware CAV operations

estimated location, size, and speed of surrounding vehicles [11], [12], and it can include the intentions of following vehicles. Existing literature does not consider the intent of the following vehicle for the left-turning movement [13]–[16]. A few studies have discussed the aggressiveness of AVs with AVs showing human-like aggressive behaviors and found that the imitation of human driving behavior could improve traffic operation and safety [11], [17]. However, existing left-turning CAV controllers do not assess the following vehicle's intent. Based on a 28-month Autonomous Vehicle Disengagement Reports Database (September 2014-January 2017), 89% of the total crashes for Autonomous Vehicles (AVs) occurred at intersections, 69% of the total crashes occurred with AV speed less than 2.24 ms-1 (5mph), and 58% of the crashes were rear-end caused by human drivers following an AV [9]. A 2016 survey found that 37% of Americans among 2,264 participants were concerned about the interaction of AVs and non-AVs [18].

In this research, we specifically focus on scenarios in an urban Transportation Cyber-Physical Systems (TCPS) environment where CAVs operate in the mixed traffic stream, as shown in Fig. 1(a). In an urban TCPS, the physical components include CAV sensors and actuators, traffic signal controllers, roadside units, and video cameras [19]–[21]. The cyber components include wireless communication, CAV controller software, and computing software in the roadside unit. Based on the in-vehicle sensor captured data about the surrounding environment, the CAV controller manages the CAV movement [22]. The objective of this research is to develop and evaluate a situation-aware CAV controller module, which will operate in response to an aggressive human driver and consider the intent of aggressiveness in the CAV decision-making controller module. As shown in Fig. 1(a), the contribution of this study is the development of a situation-aware CAV controller module, which considers the following human-driven vehicle's intent while making a left-turn at an intersection to minimize abrupt braking, and/or to minimize the waiting time of the following human-driven vehicles. The situation-aware CAV operation will be influenced by external factors (e.g., congested traffic conditions, extreme weather, etc.), and the integration of an Internet of Things (IoT) environment can assist the CAV operation in these events. Fig. 1(b) shows the interconnection of situation-aware CAV operation with other CPSs for an IoT environment, where data are exchanged, processed, and stored following a TCPS-incorporated IoT architecture [23], [24]. IoT connects different TCPS components, such as vehicles, people, and transportation infrastructure at different layers, with other CPSs, such as weather management CPS and emergency management CPS, to support the situation-aware CAV operations. For example, our situation-aware CAV operations will be affected by the roadway weather conditions at the intersection which will require the IoT services to provide real-time coordination between TCPS and weather management CPS (shown in Fig. 1(b)). Also, emergency management CPS can be activated if any collision happens due to the interaction between the aggressive driver and CAV in the IoT-enabled scenario. Inside TCPS, other systems, such as intersection management CPS, can simultaneously operate with the situation-aware CAV CPS. With the IoT integration, once the presence of an aggressive vehicle is detected by a CAV, the information can be shared with the intersection management CPS controller and other CAVs in the surrounding areas with the available communication options. The arrival of the aggressive human driver can trigger sufficient green time allocation at the traffic signals for a specific approach, to avoid any unwanted conflict with vehicles coming to the intersection from any other approaches.

This research focuses on developing a situation-aware CAV controller module that will enable safe and efficient left-turns at an intersection considering the following vehicle's aggressiveness. The controller module avoids any abrupt braking incidents and minimizes the intersection wait time of the following vehicle. Situation-aware CAVs dynamically identify the intent of the following vehicle using sensor captured data and adjust speed in real-time to reach the intersection. A video camera at the intersection will monitor the opposite through traffic stream, and using Vehicle-to-Infrastructure or V2I communication, the information will be communicated to the CAVs. In the future, when all vehicles will be connected, the gap information can be derived from the connected vehicle data using Vehicle-to-Vehicle communication. CAVs will identify the appropriate gaps in the opposite through traffic stream and accelerate/decelerate to reach the intersection to capture the appropriate gaps and clear the shared lane, which will be used by the following vehicle to move in through direction and clear the intersection.

The following Section II discusses related studies of aggressive driver identification, rear-end collision minimization, and situation-aware CAVs. Section III discusses the situation-aware CAV operation in an urban TCPS environment. Base AV operations, without considering the following vehicle's intent, are discussed in Section IV. Sections V and VI discuss the evaluation scenario and findings from this research. Finally, Section VII elaborates on the conclusions and future research.

## II. Related Study

The following subsections discuss the related studies about rear-end collision mitigation approaches, situation-aware CAVs, and driver aggressiveness. Literature related to the rear-end collision avoidance strategies and situation-aware CAVs emphasizes the fact that existing left-turning AV controllers still lack a mechanism to avoid collisions based on the following human driver's intent. While developing such a CAV controller module, we used existing research on driver aggressiveness to identify which parameters should be used by a CAV controller to detect aggressiveness of the following vehicle.

### A. Rear-end Collision Mitigation

The sudden brake by following aggressive human drivers can increase the likelihood of rear-end crashes. The aggressive driving behavior (i.e., speeding) was the contributing factor in 26% of all traffic fatalities in 2017 [25]. For autonomous vehicles, based on the 28-months Autonomous Vehicle Disengagement Reports Database (September 2014-January 2017), 58% of the crashes were rear-ended, where the following vehicles were human-driven [9]. In one study, the authors found tactile and audible collision warning systems can reduce the rear-end collision events for human drivers by increasing the brake response time, while the drivers were engaged in a cell phone conversation [26]. In a similar study, to identify the rear-end collision mitigation method for human drivers, the authors found that the audio and visual warning assisted to release the accelerator faster by the human drivers to avoid a potential rear-end crash [27]. Due to the faster accelerator release response, drivers could apply brakes gradually to avoid a collision. Another rear-end collision mitigation system for human drivers was the use of a green signal countdown timer, which was found to reduce rear-end crashes during the yellow interval [28]. The rear-end collision anticipation warning can be provided using vehicle-to-vehicle communication. As the rear-end collision avoidance application needs to satisfy strict delay constraints, the authors in one study developed a rear-end collision avoidance strategy using IEEE 802.11 standard and multi-hop broadcast system [29]. Using simulated single-lane and multi-lane scenarios, the rear-end crash avoidance strategy reduced almost all rear-end crashes for the following vehicles. AVs still lacks a mechanism to avoid rear-end crashes when the following vehicle is a human driver [9]. In this study, we have developed such a control module for left-turning autonomous vehicles to reduce the rear-end crash possibility and reduce road-rage events.

### B. Situation-aware CAV

Earlier research developed the situation-awareness for AVs based on the Partially Observed Markov Decision Process, where an autonomous agent chooses a policy for taking an action, without knowing the system state, to maximize rewards [30]. The authors considered intention recognition and sensing uncertainties in the framework and measured the conflicting vehicle intention with respect to speed. Compared to the reactive approach, the situation-aware autonomous vehicle showed fewer failure rates in different scenarios (such as interacting at roadways with T-intersection and roundabout), meaning autonomous vehicles did not always purposefully give way to the conflicting vehicles. Rather, autonomous vehicles acted proactively to reduce the waiting time. A similar method was used in another research, where the authors used four parameters (i.e., distance to the intersection, yaw rate, speed, and acceleration) to identify each vehicle's intent at an unsignalized intersection [31]. The reward function includes reward due to adherence to the traffic law, reduction in travel time and improvement in safety. Using Prescan software, the autonomous vehicles were modeled and using a driving simulator, research participants drove the human-driven vehicles. The analysis showed that, without considering the human intention, the autonomous vehicles were confused about whether to cross the intersection. In another study, the authors discussed the use of temporal domain prediction instead of spatial domain prediction to predict uncertainty in other agent's intent [32]. For autonomous agents, the authors showed that the required time to reach a destination, and maneuvering time can be designed as a Gaussian distribution. With Monte Carlo simulations, the authors demonstrated that the autonomous vehicle can safely maneuver through roundabouts while considering other vehicle's predicted position in future times. In order to reduce conflicts among multiple agents, one study investigated the empathic autonomous agent which made decisions based on a utility function (this function depends on the acceptability of any action by all agents, based on the action's future consequences) of everyone in the driving environment [33]. Here, the empathic autonomous agent made the decision that was acceptable to everyone. In another study, autonomous vehicles considered the yielding intent of merging vehicles on the freeway entrance ramp [34]. Using the acceleration value of the merging vehicle, the intent of the merging vehicle was recognized. Upon recognizing the intent, an autonomous vehicle would generate candidate strategies to minimize a cost function, which avoids conflict, passenger discomfort, excess fuel consumption, and undesirable operational outcomes. If the merging vehicles did not show the intent to yield, autonomous vehicles would slow down to avoid conflict. In this research, we have developed a situation-aware CAV controller module for one of the most critical interactions between CAVs and following aggressive vehicles, which results in the most prominent crash type, i.e., rear-end crash, for real-life autonomous vehicles [9].

*C. Driver Aggressiveness Identification*

The aggressive driver behavior was previously studied using data from the smartphone, where the authors identified the acceleration behavior of both aggressive and non-aggressive drivers to provide feedback in real-time to the corresponding drivers about their driving behavior [35]. The types of aggressive behavior included excess speeding, abrupt braking, lane changes, and aggressive U-turns. The authors in [35] considered the driver experience and road surface condition to identify the boundary values of acceptable longitudinal and lateral accelerations. The smartphone-based GPS sensor was used to obtain the real-time acceleration rate of the vehicle, and when the acceleration exceeds the allowable threshold, drivers can be alerted about their aggressive behavior in real-time. In another study, the aggressiveness behavior of a subject vehicle was identified based on the vehicle's current lane deviation possibility, speed and estimated collision time with the front vehicle [36]. The authors used both an in-vehicle sensor and camera sensor to collect the required data and trained a machine learning-based classifier (i.e., support vector machine) to identify aggressive driving behavior. The machine learning-based classifier achieved 93% accuracy to classify drivers according to their aggressive driving behavior. Vehicle trajectory data was used in another study, where the authors used relative speed, average speed, distance to leading vehicles, longitudinal jerk and lane change data from the I80 corridors in California to identify driving behavior of a subject vehicle [37]. The authors interviewed 100 participants (whose driving data were not included in the I80 database) to identify the driving behavior and level of attentiveness of the subject vehicle driver. The driving behavior identification module was incorporated into a simulated vehicle navigation system to ensure safe navigation. Both speed and lateral and longitudinal acceleration were used to derive the mathematical model of driver aggressiveness in another study, where the authors used real-world data from vehicles [38]. The authors developed a classifier using Gaussian Mixture Models and maximum-likelihood, which achieved a 92% accuracy to identify each driver's behavior. In another study, the authors used acceleration and speed of the leading vehicle, and the time gap between the leading vehicle and following vehicle to cluster different driving behaviors [39]. Based on the driving behavior and acceleration of the leading vehicle, the car-following behavior was found to be linearly stable. Vehicle data from the I80 corridors in California, available via the Next Generation Simulation database, were used to develop the car-following model. In this research, the driver's intent of the following vehicle needs to be identified. In an urban TCPS, with the following vehicle's acceleration/deceleration rate [34], [38], [39], we can directly estimate whether the following vehicle will slow down while following the leader left-turning CAV. However, time headway is another important parameter [39], as with time headway we can monitor how closely the following vehicle is following the leader CAV in the urban area. A closely-following vehicle is considered to be more aggressive, compared to the following vehicle maintaining a high headway. Thus we have used both the acceleration of the following vehicle and time headway between the subject vehicle and following vehicle to identify the following vehicle's aggressiveness in this study.

### III. SITUATION-AWARE LEFT-TURNING CAV OPERATION

Steps associated with the situation-aware left-turning CAV operation, as shown in Fig. 2(a), are: (A) detect the following vehicle's intent, (B) predict future traffic state of the opposite traffic lane, (C) identify a gap in the opposite traffic, and (D) optimize CAV movement. In this paper, the speed profile of the turning CAV is modeled following an earlier study [13]. Other components of the framework (i.e., steps (A), (B), (C), and rest part of step (D) including the inflow and outflow optimization)

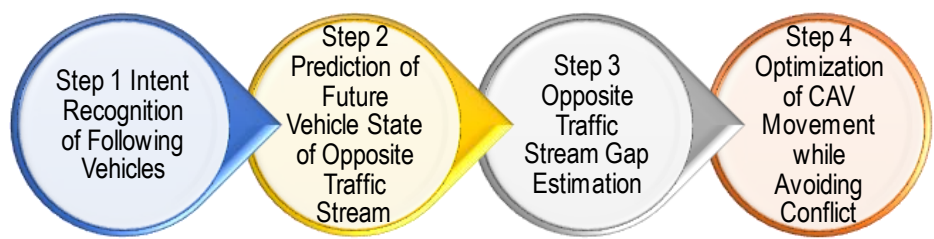

(a) Steps for the CAV left turn decision module

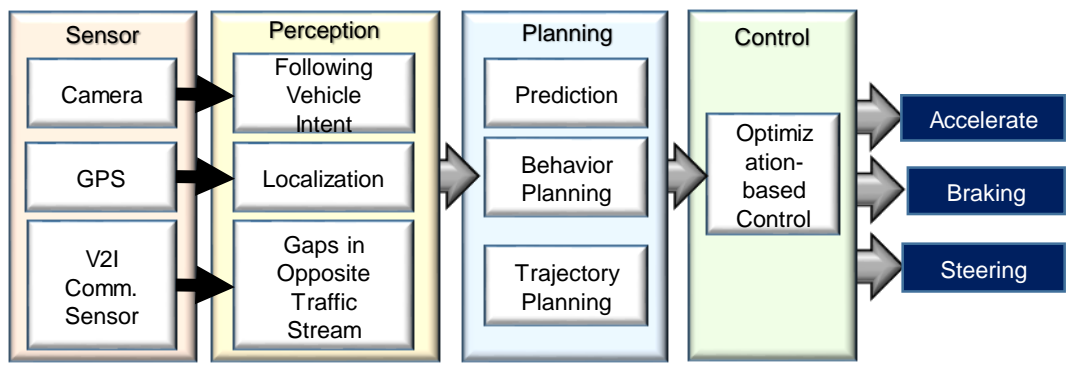

(b) Situation-aware left-turning CAV module components
Fig. 2. CAV left turn steps and decision module

are developed by us, the authors of this study. The situation-aware left-turning CAV operates depending on the surrounding situation, which for this research is an aggressive behavior of the following vehicle. If the following vehicle's intent is identified, CAV can operate accordingly to prevent or minimize negative consequences, which include abrupt hard braking, and increased waiting time for the following vehicle. Predicting the future condition of the surrounding traffic helps to take proper actions by an autonomous vehicle [40]–[42]. In this case, the opposing traffic stream's future condition will dictate the availability of the target gap at the intersection when the CAV will arrive at the intersection stop bar to initiate the left-turn maneuver [43]. If an inaccurate prediction happens while identifying gap in the opposite traffic stream, the in-vehicle sensors (camera, lidar) of the CAV will still be able to detect the incoming vehicles from the opposite direction after reaching the intersection stop bar, and the CAV will not enter the intersection. Such a case can lead to the following aggressive human driver applying sudden brake as the CAV will keep waiting at the intersection. Finally, while taking the left turn, the CAV needs to confirm that adequate gaps are there so that there will be no direct conflict with the opposite through traffic stream and the subject CAV. The four steps, as shown in Fig. 2(a), are discussed in the later subsections.

Fig. 2(b) shows the components for the situation-aware control module for left-turning CAVs. The sensors used by this module include a rear-view camera, a GPS sensor, and a V2I communication radio. Earlier studies found that CAV operations can be improved if external data can be utilized through V2X wireless communication [19], [44]. Using these sensors, the intent of the following vehicle (from the rear-view camera) and gaps in the opposite through traffic stream (from V2I communication radio using the analyzed intersection video feed from the roadside unit) are identified. If all vehicles are connected, the gap information can be derived from the connected vehicle data using Vehicle-to-Vehicle communication without any need for cameras installed at an intersection. The GPS sensor is used to identify the location of the CAV. Using the rear-view camera, the relative position of the following vehicle is identified. Based on the relative position of the following aggressive vehicle and the CAV's own position, the position of the following vehicle is identified. The planning sub-module predicts the future possible gap in the opposite through traffic stream and identifies how CAV should operate in terms of a left turn at an intersection based on the existing road traffic conditions, and traffic signal status, while also considering the speed limit of major and minor streets. This sub-module identifies the final speed to be achieved by a CAV to reach and clear the intersection. Based on the criteria identified by the planning sub-module, the control sub-module runs the optimization to estimate the speed profile to be followed by the CAV.

### A. Intent Recognition of Following Vehicles

As discussed earlier, the following vehicle can show either aggressive or non-aggressive behavior. To identify the intent, CAVs can consider the data regarding the following vehicle captured by its sensors. Different sensors can be used to obtain data from the following vehicles, and different types of data can be used. These sensors include radar, camera, and LIDAR [45]. In this research, we have considered the following vehicle's acceleration, and time headway between the CAV and the following vehicle to identify the intent of the following vehicle. The CAV follows a decision-making framework, shown in Fig. 3(a), for its left-turn maneuver. At first, it detects if there is any following vehicle. If the following vehicle is present, the CAV

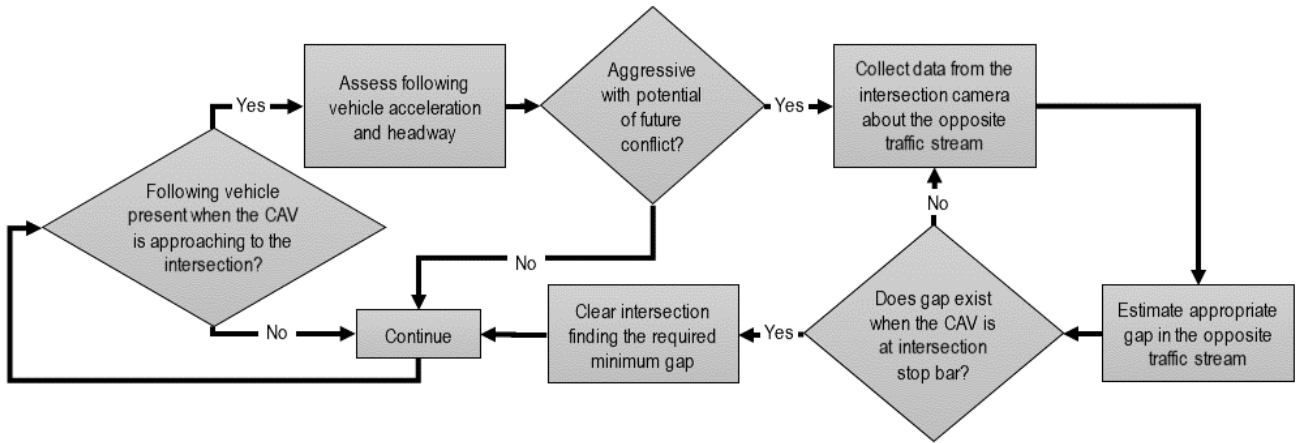

(a) Intent recognition framework

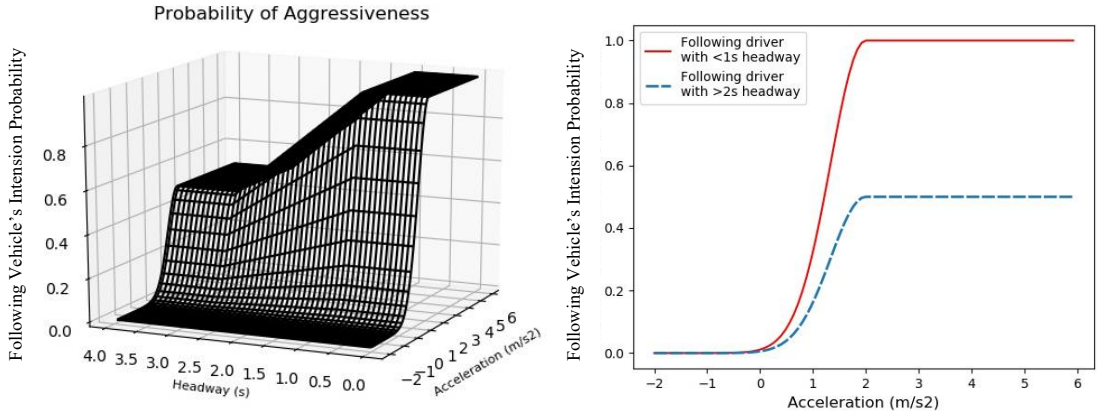

(b) Probability of aggressiveness based on acceleration and headway

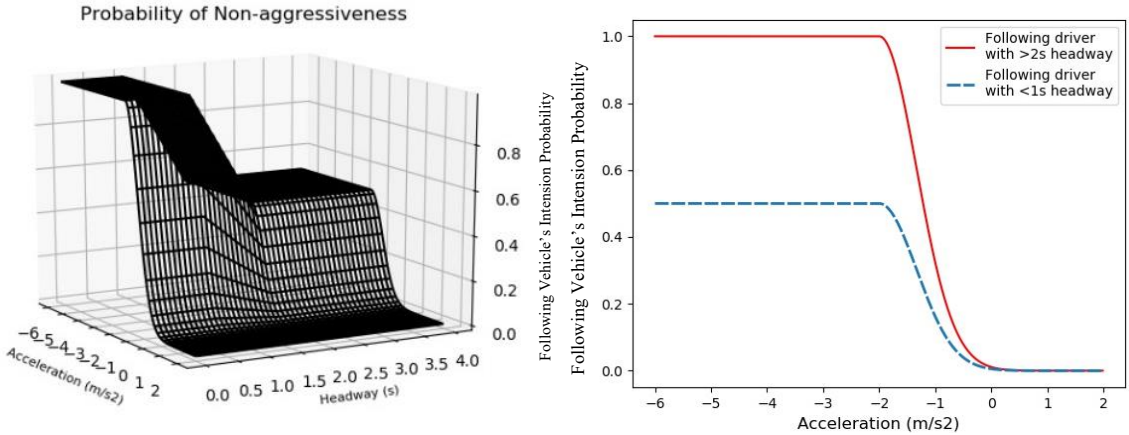

(c) Probability of non-aggressiveness based on acceleration and headway

Fig. 3. Intent recognition framework and the probability of intent

sensor captures the data of the relative position of the following vehicle $\Delta p_{t1}$ at time $t_1$. Based on its own position $p_{t_1}$ (captured by the GPS sensor available in the CAV), and $\Delta p_{t1}$, the position of the following vehicle $p_{follow,t1}$ can be estimated using (1). Using the following vehicle's position for two consecutive times, $t_1$ and $t_2$, the speed $v_{t_2}$ at time $t_2$ can be estimated using (2). From the $v_{t_2}$ (calculated speed) and $\Delta p_{t1}$ (relative position of the following vehicle at time $t_2$), the acceleration $a_{t_2}$ and time headway $th_{t_2}$ of the following vehicle at time $t_2$ can be estimated using (3), and (4), correspondingly, as shown here

$$p_{follow,t1} = p_{t1} + \Delta p_{t1} \quad (1)$$
$$v_{t_2} = \frac{p_{follow,t2} - p_{follow,t1}}{t_2 - t_1} \quad (2)$$
$$a_{t_2} = \frac{v_{t2} - v_{t1}}{t_2 - t_1} \quad (3)$$
$$th_{t_2} = \frac{\Delta p_{t2}}{v_{t_2}} \quad (4)$$

To identify the probability of the following vehicle's intent, we have used Bayes Theorem [46]. Equations ((5), and (6)) can be used to derive the probability of aggressiveness (*A*) or non-aggressiveness (*NA*) based on the attitude (*Att*) of the following vehicle, as shown

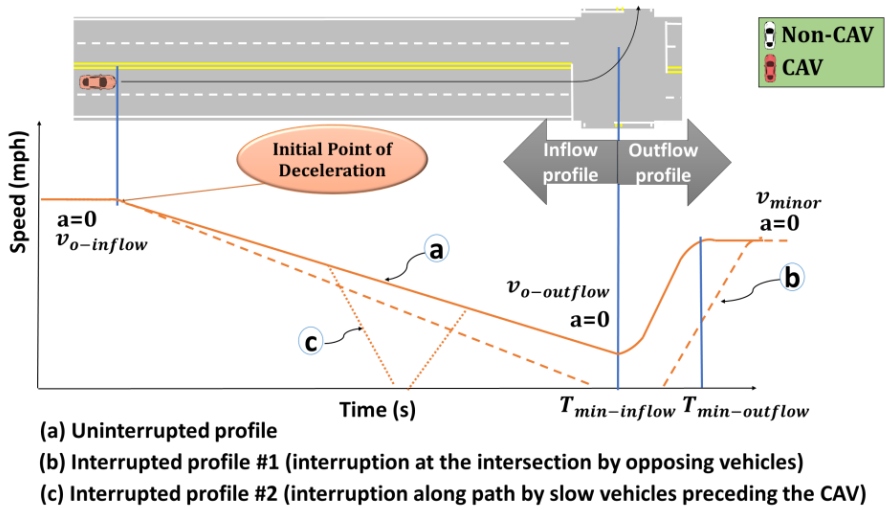

(a) Speed profiles for CAV making left-turn

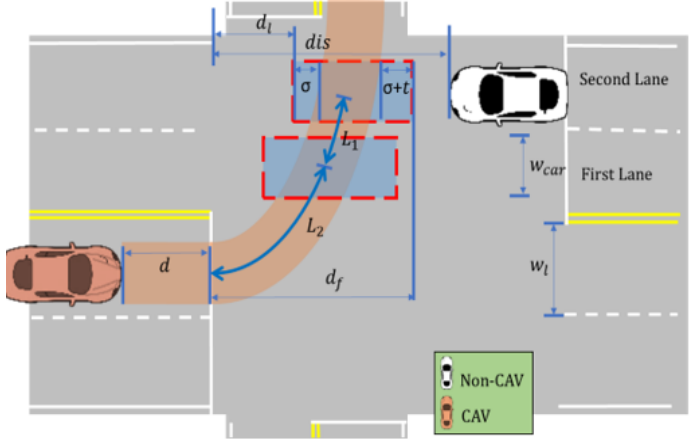

(b) Conflict area for left-turn maneuver

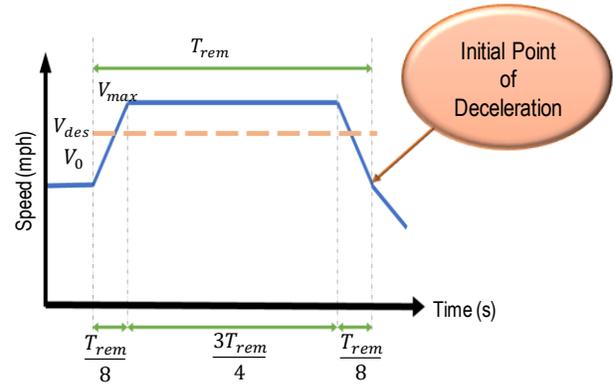

(c) CAV speed adjustment to reach the initial point of deceleration

Fig. 4. Movement of CAV making a left-turn

$$P(A|Att) = \frac{Pr\ (Att|A)\ P(A)}{P(Att|A)\ P(A) + P(Att|NA)\ P(NA)} \quad (5)$$

$$P(NA|Att) = \frac{P(Att|NA)\ P(NA)}{P(Att|A)\ P(A) + P(Att|NA)\ P(NA)} \quad (6)$$

The assumption is that there is an equal amount of chance for the following vehicle to be aggressive or non-aggressive. Thus $P(A)$ and $P(NA)$ is equal to 0.5. In order to get the $P(A|Att)$ and $P(NA|Att)$, we have considered that the aggressive and non-aggressive behaviors follow the distribution as shown in Fig. 3(b) and Fig. 3(c), correspondingly. Studies conducted on urban arterials were reviewed to obtain the threshold values for both acceleration and time headway [47], [48]. In an urban area, $2ms^{-2}$ acceleration is considered to be aggressive [48]. This value is considered as the mean of the Gaussian distribution and the standard deviation is considered to be $\frac{4}{3}ms^{-2}$ (when the acceleration is less than $2ms^{-2}$). Beyond the mean acceleration, the following vehicle will always be considered aggressive. As the CAV will have to decelerate, the following vehicle should slow down, and the non-aggressive behavior would imply that the following vehicle is slowing down. Thus, the distribution with the mean deceleration of $-2ms^{-2}$ and the standard deviation of $\frac{4}{3}ms^{-2}$ is considered as non-aggressive (when the deceleration is higher than the mean). With deceleration less than $-2ms^{-2}$, the following vehicle will always be considered non-aggressive. For time headway, a 1 second time headway is considered to be the mean of aggressive behavior, while a 2 seconds time headway represents a safe or non-aggressive behavior [47].

B. *Prediction of Future Vehicle State of Opposite Traffic Stream*

Once a CAV identifies the intent of the following vehicle, it will look for an appropriate gap in the opposite through traffic stream. The information about the opposite through traffic stream will be provided by the connected roadside unit, installed at the intersection via Dedicated Short-Range Communication, or DSRC. We have assumed a perfect communication channel for V2I communication in this study. This analysis can be extended to include communication delay and reliability following previous research on this topic. For vehicular communication, there are several options such as DSRC, 5G, LTE, and WiFi [19]. DSRC is a popular communication option as it has a dedicated spectrum for vehicular communication. From previous studies on the performance of DSRC for vehicular networking, it has been observed that DSRC has very low communication delay (~2ms) and high reliability within a short-range (~300m), which is sufficient for covering any size of signalized intersection with line-of-sight conditions [49]–[51]. For non-line-of-sight (NLOS) conditions, DSRC suffers from high path loss because of its high operating frequency (~5.9 GHz). For longer range and NLOS conditions, LTE offers a better alternative of increased coverage [50], but has a higher delay than DSRC. 5G is an emerging technology that offers a use case called ultra-reliable low-latency communication (URLLC), which would be appropriate for V2V and V2I applications [52], [53]. However, it is difficult to ensure communication reliability using one communication method. Heterogeneous wireless networking (HetNet) offers a solution to this problem, in which a CV automatically scans for available networking resources and performs horizontal or vertical handover when one communication channel is not available [54].

A camera installed at the intersection can be used to identify the gaps in the opposite through traffic and send them to the connected roadside unit for it to transmit to CAVs. In the future, when all vehicles will be connected, the gap information can be derived from the connected vehicle data using Vehicle-to-Vehicle communication. The roadside unit estimates the opposite traffic arrival time assuming that the opposite through traffic stream will maintain a constant speed while reaching the intersection. However, this assumption is not valid where human drivers can take different actions (i.e., accelerate, lane change) at any given time when they are close to the intersection. Thus, when the CAV is at the intersection with an intent to initiate the left-turn, the gap may not be there. To ensure a safe left-turn, the CAV will assess the intersection condition after reaching the intersection, in real-time, and make a final decision whether to turn left based on the data from the CAV's cameras about the approaching opposite through traffic stream. Whenever the required gap is available, the CAV will initiate the left turn to safely cross the intersection and clear the path for the following vehicle.

*C. Opposite Traffic Stream Gap Estimation*

While taking a left-turn manoeuver, two scenarios may exist. In the first one CAVs may not need to stop after reaching the intersection if there is a gap in the opposite through traffic stream right at that moment. CAV can take a left-turn without conflicting with any other vehicle after reaching the intersection at a minimum speed. This scenario can be handled by the CAV uninterrupted inflow and outflow speed profile, as shown in Fig. 4(a). In the second scenario, the interrupted inflow and outflow speed profile will be active as the CAV will have to stop at the intersection due to the presence of approaching vehicles in the opposing through traffic stream. CAV will wait for the required gap to make a left-turn based on the arriving pattern of the opposing through vehicles and start the left-turn right away when the required gap is available.

For any two way corridor with 'm' number of opposite lanes and 'n' number of vehicles on the opposing lanes at a certain time period, we have defined the vehicle sets based on the vehicles' current lane and state (App for vehicles approaching the conflict area, and Pass or P for vehicles that will pass the conflict area). $N_{A, P}$ and $N_{B, P}$ are the sets of opposing through vehicles in lane 1 and 2, respectfully, that will pass the conflict area when CAV will reach the conflict area. $N_{A, App}$ and $N_{B, App}$ are the sets of opposing through vehicles in lane 1 and 2, respectfully, that will approach the conflict area when CAV will reach the conflict area.

Fig. 4(b) shows the conflict areas at the intersection for the left-turn maneuver with red bounding boxes. We assume that the CAV will follow a parabolic path while taking the left turn at the intersection [55], [56]. For a typical two-lane corridor, the distance to the conflict area of the opposite first lane from the intersection stop line is $L_2$, and the distance from the intersection stop line to the conflict area of the opposite first lane is $(L_1 + L_2)$, as shown in Fig. 4(b). These distances can be computed with the Arc Length (*AL*) equation of the parabolic path, as shown in (7).

$$AL = \frac{1}{2}\sqrt{b^2 + 16a^2} + \frac{b^2}{8a}\ln\left(\frac{4a+\sqrt{b^2+16a^2}}{b}\right) \quad (7)$$

For a left-turn parabolic path at the intersection, the length along the parabola axis (*a*) and perpendicular chord length (*b*) can be calculated as $a = 2.5w_l$ and $b = 3w_l$ for a typical corridor with two lanes in each direction of the major road. $w_l$ is the lane width, and $w_{car}$ is the vehicle width. Here $d_l$ is the distance between the CAV direction stop line and the end of the conflict area. We have defined $d_f$ as the distance between the stop line at lanes from which a CAV will start the left turn maneuver and the start of the conflict area. We have defined a distance threshold for both sides of the conflict area compared to the parabolic path of the CAV. For the start and end of the conflict points, the distance thresholds beyond the CAV's projected path are $\sigma$ and $\sigma+t$, as shown in Fig. 4(b). Considering a two-lane-two-way corridor, both $d_l$ and $d_f$ from Fig. 4(b) for any CAV *i* can be calculated from the following (8), to (11). Here the first lane means the closest opposite lane for the left-turning CAV, and the second lane means the farthest opposite lane, as shown in Fig. 4(b).

$$d_{l-second\ lane,i} = 1.41\sqrt{w_l} - \frac{w_{car,i}}{2} - \sigma, i\epsilon N_{B, P} \quad (8)$$

$$d_{f-second\ lane,i} = 1.41\sqrt{w_l} + \frac{w_{car,i}}{2} + \sigma+t, i\epsilon N_{B, App} \quad (9)$$

$$d_{l-first\ lane,i} = \sqrt{w_l} - \frac{w_{car,i}}{2} - \sigma, i\epsilon N_{A, P} \quad (10)$$

$$d_{f-first\ lane,i} = \sqrt{w_l} + \frac{w_{car,i}}{2} + \sigma+t, i\epsilon N_{A, App} \quad (11)$$

*D. Optimization of CAV Movement while Avoiding Conflict*

Once the following vehicle intention is known and gaps from the opposite through traffic stream are identified, the CAV controller module needs to estimate its speed profile for the remaining distance. The CAV will follow the speed profile, shown in Fig. 4(a), to clear the path for the following vehicles, or at least to minimize the waiting time for the following aggressive vehicle. The speed of the turning vehicle can be modeled as a function of time with the polynomial of third-degree as discussed in [13]. The slope of a speed profile means acceleration, and the slope of the acceleration profile is called a jerk. For an initial time, $t_o$ we express the speed, acceleration and jerk values as $v_o$, $a_o$ and $J_o$. We have defined the slopes of the vehicle jerk as $\acute{j}$. The value of $a_o$ is zero. For any time $t$, the jerk, acceleration and speed can be calculated using the following (12), (13), and (14), respectively. The same equations can be applied to both inflow and outflow speed profiles.

$$J_t = J_o + \acute{j}t \tag{12}$$
$$a_t = a_o + J_0 t + \frac{1}{2}\acute{j}t^2 \tag{13}$$
$$v_t = v_o + a_0 t + \frac{1}{2}J_0 t^2 + \frac{1}{6}\acute{j}t^3 \tag{14}$$

To get the optimal speed profile, the optimization is computed in two steps. In the first step, the inflow speed profile is optimized using the optimized $\acute{j}$ for the inflow. We minimize jerk at which the CAV will reach the intersection (i.e., $J_{T_{min-inflow}}$). In the second step, based on the output of the inflow optimization model, the outflow speed profile is optimized. Here we minimize the jerk at which the CAV will enter the minor street (i.e., $J_{T_{min-outflow}}$). The input of the optimization model is the initial inflow speed, $v_{o-inflow}$. The speed at which the CAV will reach the intersection needs to be close to zero, so the target speed range is considered to be within 0.1 $ms^{-1}$ to 2.5 $ms^{-1}$. $a_{T_{min-inflow}}$ is the acceleration of CAV after reaching the intersection, which is 0. The initial jerk ($J_{o-inflow}$) at the initial point of deceleration is confined within the boundary of 1.5$ms^{-3}$, as that is defined as the limit of the comfortable jerk [57]. The boundary values for the slope of jerk ($\acute{j}_{inflow}$) are derived from [13]. The optimization objective, constraints and decision variables are given below.

Optimization objective for inflow:
$$min\ (J_{T_{min-inflow}}) \tag{15}$$

Subject to,
$0.1\ ms^{-1} < v_{T_{min-inflow}} < 2.5\ ms^{-1}$
$-1.5\ ms^{-3} < J_{o-inflow} < 1.5\ ms^{-3}$
$0.1\ ms^{-4} < \acute{j}_{inflow} < 0.8\ ms^{-4}$
$0\ s < T_{min-inflow} < T_{min-inflow,max}$
$a_{T_{min-inflow}} = 0$

Decision variables,
$\acute{j}_{inflow}, J_{o-inflow}, v_{T_{min-inflow}}, T_{min-inflow}$

The maximum available time to reach the intersection stop line ($T_{min-inflow,max}$) can vary based on the traffic conditions and geometric characteristics of the corridor. Once the desired target speed ($v_{T_{min-inflow}}$) from the initial optimization is available, the second optimization is conducted for the outflow model. For this outflow, the optimization model is provided in the following (16).

Optimization objective for outflow:
$$min\ (J_{T_{min-outflow}}) \tag{16}$$

Subject to,
$v_{T_{min-outflow,min}} < v_{T_{min-outflow}} < v_{T_{min-outflow,max}}$
$-1.5\ ms^{-3} < J_{o-outflow} < 1.5\ ms^{-3}$
$-0.2\ ms^{-4} < \acute{j}_{outflow} < -0.6\ ms^{-4}$
$5\ s < T_{min-outflow} < T_{min-outflow,max}$
$a_{T_{min-outflow}} = 0$

Decision variables,
$\acute{j}_{outflow}, J_{o-outflow}, v_{T_{min-outflow}}, T_{min-outflow}$

The maximum and minimum boundary values of the speed (after entering the side street) to be achieved by the CAVs ($v_{T_{min-outflow,min}}$ and $v_{T_{min-outflow,max}}$) and the upper limit of time ($T_{min-outflow,max}$) depend on the minor street corridor. The initial jerk ($J_{o-outflow}$) at the beginning of the lest-turn is confined within the boundary of 1.5$ms^{-3}$ [57]. The boundary values for the slope of jerk ($\acute{j}_{outflow}$) are derived from [13]. $a_{T_{min-outflow}}$ is the acceleration of CAV after entering the minor street, which is 0. Once the optimization is done, the distance required to initiate CAV deceleration to reach the intersection stop line ($d_{T_{min-inflow}}$) can be estimated with the following (17), where $d_o$ is zero.

$$d_{T_{min-inflow}} = d_o + v_0 T_{min-inflow} + \frac{1}{2} a_o T_{min-inflow}^2 + \frac{1}{6} J_0 T_{min-inflow}^3 + \frac{1}{24} \acute{j} T_{min-inflow}^4 \tag{17}$$

The point from which the CAV needs to slow down is shown as 'Initial Point of Deceleration' in Fig. 4(a). The distance between this initial point of deceleration and the intersection stop line is $d_{T_{min-inflow}}$. To reach the initial point of deceleration, the CAV adjusts its speed to reach the point soon. The desired speed ($v_{des}$) to reach the slow down point can be calculated simply by dividing the current distance from the CAV to the initial point of deceleration with the available time. However, $v_{max}$ is calculated to create a trapezoidal shape so that the CAV can smoothly increase its speed and slow down, as shown in Fig. 4(c). The CAV chooses the appropriate gap which it can utilize so that the speed to reach the initial point of deceleration, $v_{max}$ does not exceed the speed threshold (i.e., speed limit + 2.24 ms$^{-1}$ (5 mph)).

## IV. BASE LEFT-TURNING AV OPERATION

Traditionally, the AV does not have any communication capabilities and it does not have to consider the following vehicle's intent (to yield or not yield) to make a left turn at the intersection. The AV uses the front camera to detect the opposite approaching vehicle, and based on the distance between the AV and the opposing vehicles, the AV calculates the gap and evaluates if the gap is acceptable. In a study conducted in California, the authors studied the left-turn gap acceptance value from 1573 observations [58]. For human drivers, the authors found that the 15%, 50%, and 80% of the accepted gap lengths were 4.1, 6, and 8.6 seconds, respectively. For this study, after trial-and-error with the simulated scenario, we have found 5 seconds is the accepted gap for the AV left-turn maneuver. For gaps less than 5 seconds, a collision occurs between AVs and the opposite through non-AVs. In this scenario, the following vehicle starts the journey after 8 seconds of the leader AV. In this study, two base AV operations are considered, one operating with the objective of travel time minimization, and the other following the speed limit of the road without having any performing any travel time optimization.

*A. Base AV Operation #1 without travel time minimization*

In this method, the AV does not perform any optimization, rather it strictly follows the posted speed limit of the corridor while approaching the signalized intersection. After entering the corridor, the AV speeds up to the posted speed limit, if there is no obstacle at front. AVs use in-vehicle sensors to detect

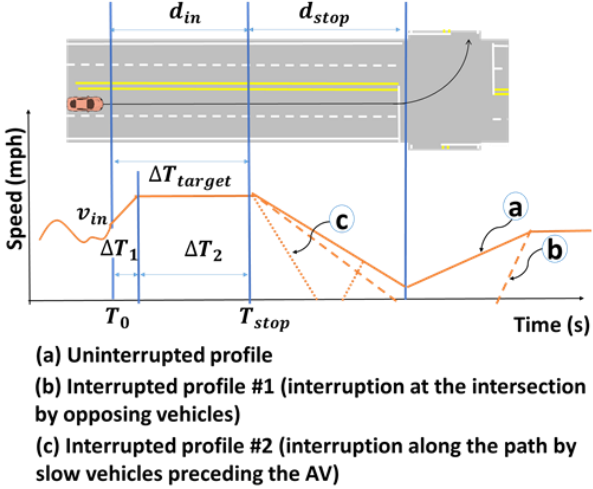

(a) Uninterrupted profile
(b) Interrupted profile #1 (interruption at the intersection by opposing vehicles)
(c) Interrupted profile #2 (interruption along the path by slow vehicles preceding the AV)

Fig. 5. Base #2 AV operation with minimized travel time

objects in the surrounding environment while approaching the intersection and to find a gap in the opposite through vehicle stream while making a left-turn.

*B. Base AV Operation #2 with travel time minimization*

We have adopted the AV operation developed by Fayazi and Vahidi (2018), where the AV travel time for a signalized corridor is optimized in a mixed traffic environment [59]. We have modified the AV operation formulation as our study is specifically for a left-turning AV. Our study assumes that AVs will strictly follow the posted speed limit while travel time is optimized. To clear the intersection area while making the left-turn, AVs solely rely on their sensors.

As shown in Fig. 5, a stopping distance ($d_{stop}$) is required by the AVs approaching the intersection to stop the vehicle if any conflicting vehicle is present at the intersection. It is calculated using (18), where $t_r$ is the reaction time of AV (i.e., 0.5 second), $v_{max}$ is the maximum posted speed to be followed by the AV and $a_{dec}$ is the deceleration of AV (i.e., $-1.5 ms^{-2}$).

$$d_{stop} = t_r v_{max} - \frac{v_{max}^2}{2a_{dec}} \quad (18)$$

While at the intersection, the AV may or may not need to stop at the intersection stop bar based on the left-turning gap availability (shown with speed profile a, and b in Fig. 5). However, the speed of AV can be influenced by the front vehicles in the same direction (shown with speed profile c in Fig. 5), and AV will need to apply emergency brake to stop and accelerate to achieve the recommended speed.

The travel time optimization formulation, as discussed in [59] reduces the travel time required to reach the initial point of $d_{stop}$, which is $T_{stop}$. At a certain timestamp $T_O$, when the AV has just entered the corridor, the following optimization initiates.

Optimization objective for AV:
$$min \ (T_{stop} - T_0) \quad (19)$$

Subject to,
$T_{stop} \geq T_0 + \Delta T_{target}$
$\geq T_0 + \Delta T_1 + \Delta T_2$
$\Delta T_2 \geq 0$
$\Delta T_1 = \frac{v_{max} - v_{in}}{a_{in}}$
$\Delta T_2 = \frac{d_{in}}{v_{max}} - \frac{v_{max}^2 - v_{in}^2}{2 a_{in} v_{max}}$
$v_{in-min} < v_{in} < v_{in-max}$
$0.5 \ ms^{-2} < a_{in} < 1.5 \ ms^{-2}$
$0 \ s < T_{stop} < T_{max}$

Decision variables,
$T_{stop}, v_{in}, a_{in}$

Here $v_{in}$ and $a_{in}$ are the initial speed and acceleration to be followed by AV. $d_{in}$ is the distance between the AV's initial position (where the optimization occurs) to the initial location of $d_{stop}$, and $\Delta T_{target}$ is the time required by the AV to cross the distance $d_{in}$. The boundary values of initial speed ($v_{in}$) and the upper limit of time ($T_{max}$) depend on the corridor. $a_{in}$ is confined to ensure the acceleration is within a comfortable range. Once the optimization is solved, the AV proceeds forward with $v_i$ and achieve $v_{max}$ by $\Delta T_1$ time interval, and maintains $v_{max}$ for the $\Delta T_2$ time interval. $\Delta T_{target}$ is the summation of $\Delta T_1$ and $\Delta T_2$. After reaching the intersection, the AV relies on the in-vehicle sensors to make the left turn. In this operation, AVs do not have any wireless connectivity with the outside world.

## V. CASE STUDY

We have evaluated the situation-aware left-turning module for CAVs using a case study within a simulated environment. The following subsections discuss the case study area, base scenario, and situation-aware CAV module.

*A. Study Area*

A case study is conducted with a simulated intersection from Perimeter Road, Clemson to evaluate the performance of the situation-aware CAV controller module. To simulate the non-CAVs of the mixed traffic stream, we have used Simulation of Urban Mobility (SUMO) software, while to simulate CAVs (including in-vehicle sensors), the following aggressive vehicle, and communication infrastructure, we have used a robot simulator, Webots [60]. The simulation parameters are listed in Table I. The aggressive human driver is simulated in a way that it will follow the posted speed limit, and will not decelerate properly following the leading CAV. It will apply hard brake only when it is very close to the leading CAV, while the CAV is waiting to make a left turn at the intersection. The major corridor of this intersection has two lanes, while the minor corridor has one lane. We have evaluated the simulated network with multiple scenarios while varying the opposite direction traffic. The traffic signal phase for the shared lane is considered to be permissive green, meaning left-turning vehicles need to wait for the appropriate gaps in the opposite through traffic stream. In this experiment, we have restricted the lane-changing capability of the following vehicle. This scenario simply means that due to the presence of heavy traffic in the same direction, the following aggressive vehicle cannot make any lane change. The author has considered 600, 800, and 1000 vehicle per hour per lane (vphpln) opposite through traffic. For the non-CAVs, the speed distribution is set up in such a way so that 95% of the vehicles drive within 70%-110% of the speed limit. The speed limit of the corridor is 13.4 ms$^{-1}$ (30 mph). The comparison of the base scenario and situation-aware CAV is conducted based on 30 simulation runs for each scenario with different approaching through traffic volume from the opposite direction.

TABLE I
CASE STUDY SIMULATION PARAMETERS

| Simulation Parameters | VALUES |
|---|---|
| Opposite through traffic (SUMO) | 600, 800 and 1000 vphpln |
| Speed distribution of opposite through vehicle (SUMO) | 95% of the vehicles drive within 70%-110% of the speed limit |
| Major road speed limit | 13.4 ms$^{-1}$ (30 mph) |
| Minor road speed limit | 7 ms-1 (15 mph) |
| Major road length | 337m (0.2 mile) |
| Intersection roadside unit (Webots) | emitter radio node, range 300m (0.19 mile) |
| CAV model (Webots) | BMWX5 Robot |
| CAV sensors (Webots) | •Back camera node (with recognition mode on, max range 200m) <br> •Distance sensor node (generic) <br> •GPS node <br> •Gyro node <br> •Receiver radio node |
| Base AV sensors (Webots) | •Front camera node (with recognition mode on, max range 200m) <br> •Distance sensor node (generic) <br> •GPS node <br> •Gyro node |

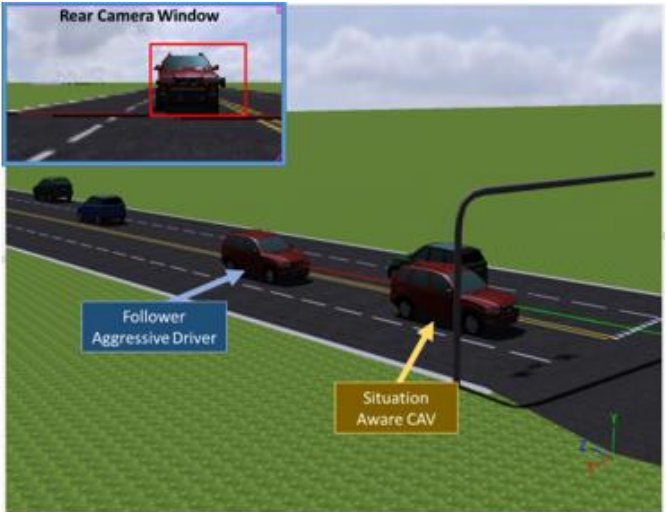

Fig. 6. Situation-aware CAV tracking following vehicle

### B. Base Scenario with Autonomous Vehicle

In both AV scenarios (operating without and with travel time minimization), the following vehicle starts the journey after 8 seconds of the leader AV. For our case study, the boundary values of speed, $v_{in-min}$ and $v_{in-max}$ in the base #2 AV operation (with travel time minimization), are considered to be 11.5 and 12.5 ms$^{-1}$, respectively. The upper limit of time ($T_{max}$) is considered to be 60s. The maximum posted speed to be followed by the AV ($v_{max}$) is 30 mph.

### C. Situation-aware CAV

The goal of the situation-aware CAV controller module is to clear the path from the shared lane for an aggressive through vehicle, so that the aggressive driver does not need to apply a hard brake. If no safe gap is available, the CAV will try to clear the path of the aggressive following vehicle by making a left-turn as soon as possible. We have considered the maximum available time to reach the intersection stop line ($T_{min-inflow,max}$) as 60 seconds for this analysis, and the maximum and minimum boundary values of the target speed (after entering the side street) to be achieved by the CAVs ($v_{T_{min-outflow,min}}$ and $v_{T_{min-outflow,max}}$) as 6 $ms^{-1}$ and 7$ms^{-1}$, respectively, based on the minor street speed limit from the study area. In this research, we have used the acceleration of the following vehicle and time headway between the CAV and following vehicle to identify the aggressiveness or non-aggressiveness of the following vehicle. The CAV uses a back view camera to capture data related to the following vehicles, using (1), to (4). The range of cameras currently used in AVs can be up to 250 meters [61]. In this research, we have considered the range to be 200 meters. As the in-vehicle camera sensor is used to capture in real-time the following vehicle's movement, there is no delay in data collection. Also, the computational delay is negligible as the CAV will detect an aggressive vehicle which is 200 meters in the upstream location, leaving CAV with sufficient time to react. Fig. 6 shows the rear camera window of a situation-aware CAV while tracking the following vehicle. Similar to the base scenario, here the following vehicle starts the journey after 8 seconds of the leader CAV. The author has used MIDACO solver to solve the optimization function in real-time [62].

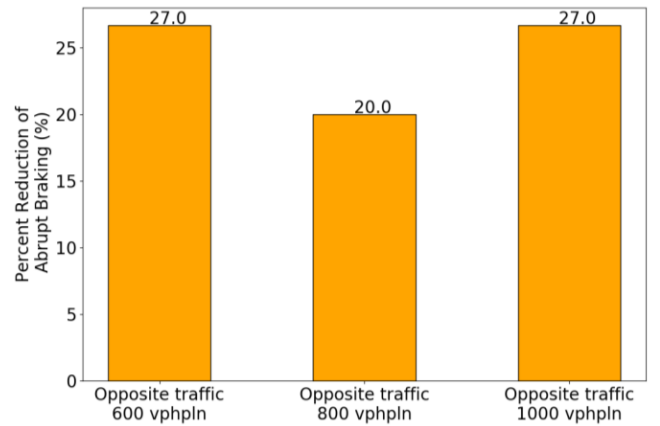

Fig. 7. Abrupt braking reduction by situation-aware CAV

The arrival time of the vehicles approaching from the opposite through needs to be estimated. Here, one assumption is that a video camera will be installed at the intersection, and it will be used to estimate arrival times of the opposite through vehicles. To identify the start and end of the conflict points in the opposing through traffic stream, $\sigma$ and $t$ values are considered to be 0.6 meter (2 ft.) and 1.2 meter (4 ft.) [63]. These small distance thresholds were considered as they provide more gaps for a CAV's left-turning maneuver that would avoid rear-end crash likelihood with a following aggressive driver. The roadside units, installed at the intersection, will share the camera captured data with the CAV using the V2I communication. The intersection video camera will use V2I communication only to share the information about the approaching through vehicle stream with the CAV. In a previous study, the authors implemented a real-world TCPS application for pedestrian movement detection using a video camera-enabled connected roadside unit [64]. The same experimental setup can be used to detect gaps between approaching vehicles on the opposing through lanes. Similar data can be captured through V2V communication if all of the approaching vehicles at the intersection are connected vehicles. After reaching the intersection, a CAV utilizes data from the intersection camera/RSU about the location of the approaching vehicles from the opposite through direction.

## VI. ANALYSIS AND FINDINGS

The following subsections discuss the findings for both leader CAV and the following vehicle.

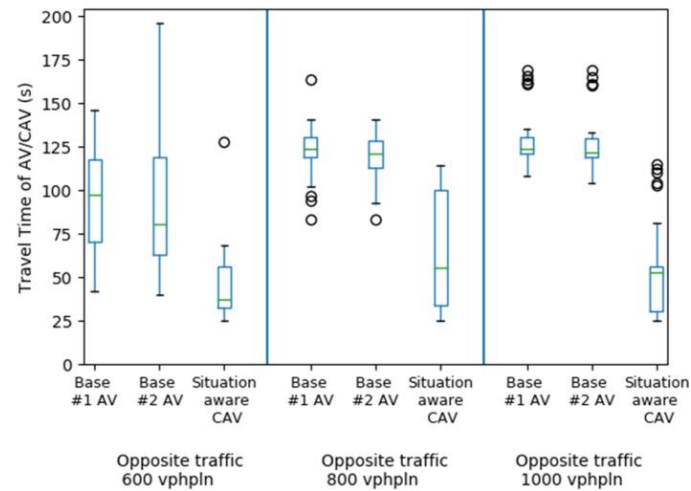
(a) Travel time variations of the leader AV/CAV

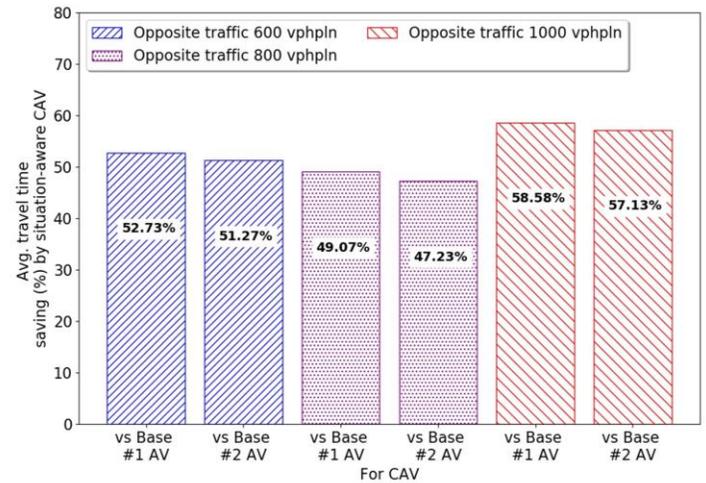
(b) Travel time saving of the CAV

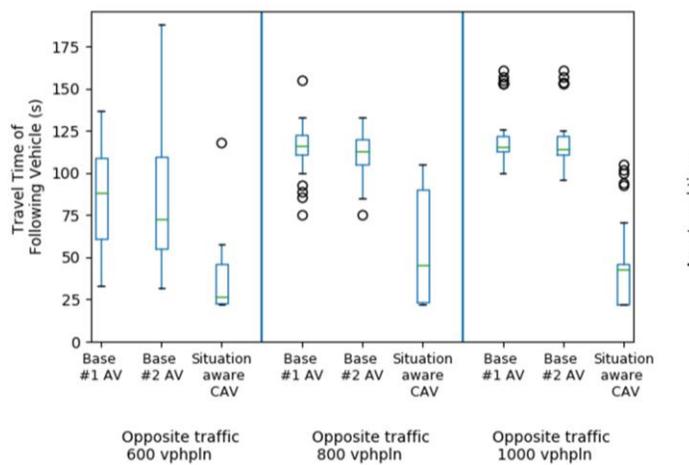
(c) Travel time variations of the vehicle following an AV\CAV

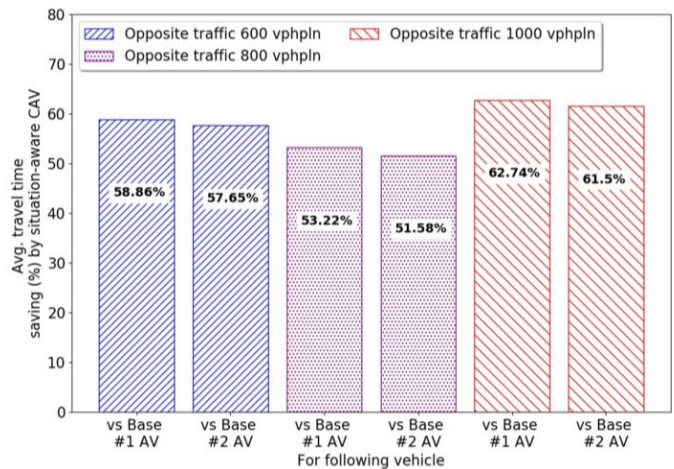
(d) Travel time saving of the vehicle following an AV\CAV

Fig. 8. Travel time Findings

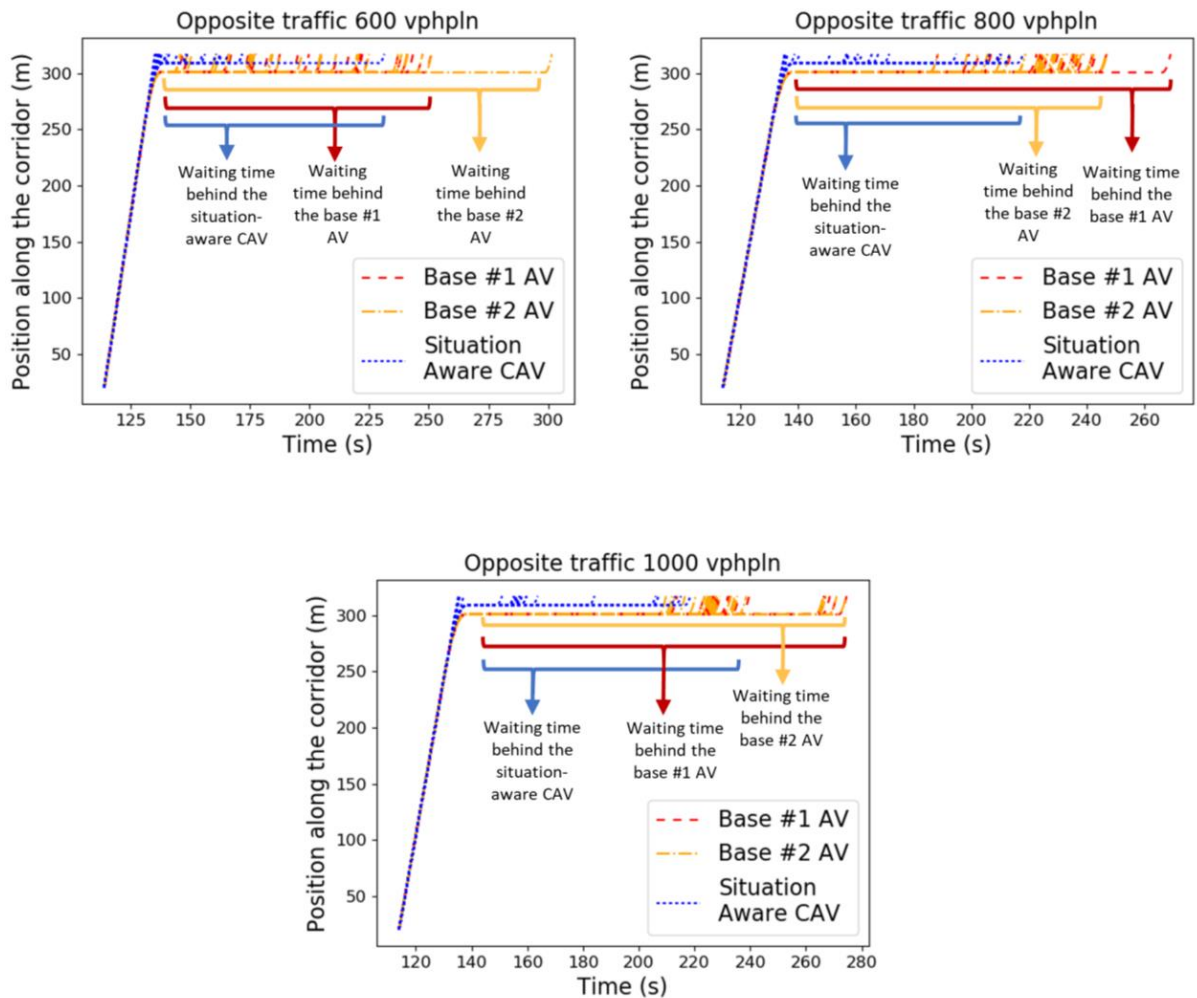

Fig. 9. Following vehicle progression

### A. Abrupt Braking of Aggressive Following Vehicle

In this study, the abrupt braking of the aggressive driver is characterized by a sudden reduction of speed. We have quantified the number of abrupt braking event reduction by the situation-aware CAVs. As shown in Fig. 7, the situation-aware CAV controller reduces 27% of the abrupt braking of the following vehicle for the 600 and 1000 vphpln opposing through traffic, compared to both base scenarios with an AV without situation-awareness (i.e., with and without travel time optimization). For 800 vphpln opposite traffic, the abrupt braking reduction decreases to 20%.

### B. Travel Time for CAV and Following Vehicle

We have estimated the travel time for the subject vehicle from the start point of the corridor from which the vehicle starts to move to the intersection to the start point of the target corridor after taking the left turn. The mean values shown in box plots in Fig. 8 (a) signifies that the Base #2 AV operation with travel time minimization objective has a lower travel time compared to the scenario without any optimization (i.e., Base #1). However, compared to both Base #1 and #2 AV operations, the situation-aware CAV controller module decreases the travel time for the vehicle itself for each scenario. Fig. 8 (b) shows the percent reduction of average travel time by the situation-aware CAV compared to both base AV scenarios (with and without travel time minimization). Compared to the base #2 AV operation (with travel time minimization), the situation-aware CAV module offers average travel time reductions of 51%, 47%, and 57%, for the 600, 800, and 1000 vphpln scenarios, respectively.

Similar results are derived by observing the following vehicle, as shown in Fig. 8 (c) and Fig. 8 (d). Examining the mean values of box plots in Fig. 8 (c), Base #2 AV operation with travel time minimization objective reduces the travel time of the following vehicle compared to the scenario without any optimization (i.e., Base #1). However, the following aggressive vehicle's travel time is further reduced by the situation-aware CAV controller module for each scenario with different opposite through traffic volumes. In the situation-aware CAV scenario, the average travel time savings for the following vehicle, compared to the base #2 AV operation (with travel time minimization), are 58%, 52%, and 62% for the 600, 800, and 1000 vphpln opposite through vehicle stream, respectively.

### C. Aggressive Following Vehicle Progression

One of the purposes of the situation-aware CAV controller module is to clear the path for the following aggressive vehicle driver so that the following vehicle does not need to wait for a long time. The progression profile (i.e., vehicle location with respect to time) of the aggressive vehicle, following a CAV, provides a clear picture of the impact of the situation-aware CAV controller module. Fig. 9 shows the progression of the vehicle following an AV/CAV in base #1 AV, base #2 AV, and situation-aware CAV scenario with time. The study corridor length is 337m, and the horizontal line after 300m in Fig. 9 means that the vehicle following an AV/CAV is stopped close to the intersection because of the front AV/CAV. The length of the horizontal lines is proportional to the waiting time of the vehicle following an AV/CAV. As shown with the blue dotted lines in Fig. 9, the waiting time of the following vehicle is the lowest while the aggressive human driver is operating behind a situation-aware CAVs (regardless of the opposing through traffic volume) compared to the vehicle following an AV (base #1 and #2 AV). It is evident from the progressions that the V2I communication enabled situation-aware CAV helps the aggressive vehicle following a CAV to quickly progress through the intersection compared to the base scenario with AV without V2I communication and situational awareness.

## VII. Conclusions

The presence of aggressive human drivers in a mixed traffic stream makes the operation of CAVs challenging, as aggressive drivers tend to follow the leader vehicle very closely. Any sudden movement change by a leader CAV has the potential to cause abrupt behavior by the following vehicle, which may result in road rage and/or a rear-end crash. Also, human drivers often could take unethical advantages of the defensive driving behavior of AVs. If CAVs can act based on surrounding situations, they can mimic human behavior more closely, which will reduce the confusion among the surrounding human drivers about any future actions of CAVs. In this study, the situation-aware CAV uses its own rear camera sensor to identify the following vehicle's intent. Once the CAV determines that the following vehicle is aggressive, it determines the appropriate gap in the opposite through traffic stream, optimizes the speed profile, and increases its speed to reach the initial point of deceleration to initiate the left-turn. If a safe gap is not available when the CAV reaches the approach intersection stop line, the CAV evaluates data from the roadside units about available gaps on the opposing through lanes and prepares to make a left-turn immediately when the required gap is available. The overall decision-making module helps to clear the intersection as soon as possible to reduce the travel time of the following aggressive vehicle.

Based on the analysis conducted in this research, we have found that the situation-aware CAV improves the operational condition compared to the base scenario with only AV (without any V2I communication) for different flow rates in the opposite through vehicle stream. The situation-aware CAV controller module reduces the number of abrupt braking by 27%, 20%, and 27% for opposing through traffic stream with 600, 800, and 1000 vphpln, respectively, compared to the base scenario without situational awareness of AVs. While assessing the travel time reduction, the situation-aware CAV scenario reduces travel time for a CAV, compared to the base scenario with AV (operating with travel time optimization), as much as 51%, 47%, and 57% for the 600, 800, and 1000 vphpln opposing through vehicles, respectively. Similar improvements are found for the following vehicles with 58%, 52%, and 62% travel time savings for the 600, 800, and 1000 vphpln opposing through vehicles, respectively.

The desired benefit may not be achieved if CAVs cannot be proactive to reduce potential conflicts due to responding to an aggressive following non-CAV. With an increasing penetration level of CAVs, a cooperative movement can be enabled with CAVs in the opposing traffic stream to help a left-turning CAV find a gap if a following aggressive vehicle is present. Also, the human driver's aggressiveness level can vary from person to person. Developing the situation-aware CAV module for a wide range of driver aggressiveness can help CAV take actions based on the characteristics of the specific following driver. A situation-aware CAV operation will be influenced by external factors (e.g., congested traffic condition, extreme weather, etc.) where front vehicles may need to do sudden lane change. Having the capability of dynamic speed profile optimization can help the CAV operation in such scenarios, which can be studied in the future. Future studies should also be conducted to evaluate the impacts of wireless communication options on the situation-aware CAV controller operation. Finally, a real-world evaluation of the situation-aware CAV controller module presented in this paper should be conducted to validate the operational benefits in real-life.

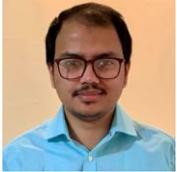
**Sakib Mahmud Khan** is the Assistant Director of Center for Connected Multimodal Mobility at Clemson Univresity. Before that, he was a post-doctoral research scholar working at California Partners for Advanced Transportation Technology (PATH), University of California Berkeley. Before joining PATH, he worked as a research specialist at the Center of Connected Multimodal Mobility at Clemson University. He received his Ph.D. and M.Sc. from Clemson University in 2019 and 2015, respectively.

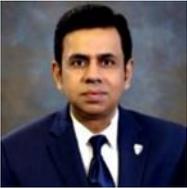
**Mashrur "Ronnie" Chowdhury** (SM' 14) is the Eugene Douglas Mays Professor of Transportation at Clemson University. He is the director of the USDOT Center for Connected Multimodal Mobility (C2M2) ((http://cecas.clemson.edu/c2m2). He is the co-director of the Complex Systems, Analytics and Visualization Institute (CSAVI) (http://clemson-csavi.org) at Clemson University. He is the director of the Transportation Cyber-Physical Systems Laboratory at Clemson University. He previously served as an elected member of the IEEE ITS Society Board of Governors and is currently a senior member of the IEEE. He is a Fellow of the American Society of Civil Engineers (ASCE) and an alumnus of the National Academy of Engineering (NAE) Frontiers of Engineering program. Dr. Chowdhury is a member of the Transportation Research Board (TRB) Committee on Artificial Intelligence and Advanced Computing Applications, and the TRB Committee on Intelligent Transportation Systems. He is the founding advisor of the "IEEE Intelligent Transportation Systems Society (ITSS) Student Chapter" at Clemson University. He is a registered professional engineer in Ohio.